%% file: main.tex
\documentclass[letterpaper, 10pt, conference]{ieeeconf}
\IEEEoverridecommandlockouts

\usepackage[utf8]{inputenc}
\usepackage{graphicx}
\usepackage{physics}
\usepackage{amsmath, amsfonts, amssymb}
\usepackage{xurl}
\usepackage[hidelinks]{hyperref}
\usepackage[dvipsnames]{xcolor}
\usepackage{cite}
\usepackage{multirow}
\graphicspath{{./figs/}}
\usepackage{tikz}
\usetikzlibrary{arrows, positioning, calc}
\DeclareMathSymbol{\shortminus}{\mathbin}{AMSa}{"39}
\DeclareMathAlphabet{\mathpzc}{OT1}{pzc}{m}{it}

\newcommand{\ril}[1]{\mathbb{R}^{#1}}
\newcommand{\ellips}[2]{\mathcal{E}(#1, #2)}

\title{Robust Adaptive Predictive Control for Hook-Based \\Aerial Transportation Between Moving Platforms}
\author{Péter Antal$^{1}$, Andrea Carron$^{2}$, Melanie Zeilinger$^{2}$, Roland Tóth$^{1,3}$, and Tamás Péni$^{1}$
\thanks{$^{1}$P. Antal, R. Tóth, and T. Péni are with the Systems and Control Lab, HUN-REN Institute for Computer Science and Control, Budapest, Hungary (email: antalpeter@sztaki.hu, peni@sztaki.hu, r.toth@tue.nl).}
\thanks{$^{2}$A. Carron and M. Zeilinger are with the Institute for Dynamic Systems and Control, ETH Zürich, Zürich, Switzerland (email: carrona@ethz.ch, mzeilinger@ethz.ch).}
\thanks{$^{3}$R. Tóth is also affiliated with the Control Systems Group of the Eindhoven University of Technology, Eindhoven, The Netherlands.}
}

\begin{document}

\maketitle

\begin{abstract}
This paper presents a novel model predictive control (MPC) approach for autonomous pick-and-place between moving platforms with a hook-equipped aerial manipulator. First, for accurate and rapid modeling of the complex dynamics, a digital twin model of the quadcopter equipped with a hook-based gripper, implemented in MuJoCo, is constructed and used as the predictive model for the MPC. To handle uncertainties of the predictive model (e.g. due to aerodynamics and uncertain payloads), a robust adaptive MPC approach is proposed. By systematic integration of zero-order robust optimization (zoRO) based uncertainty propagation and an extended Kalman filter (EKF) for parameter estimation, the MPC algorithm ensures robust constraint satisfaction, high performance, and computational efficiency. The effectiveness of the proposed method is evaluated in complex simulated scenarios and in real-world flight experiments.

Video: \url{https://youtu.be/l_L7mpUYJqU}

Code: \url{https://anonymous.4open.science/r/quadcopter-transport-mujoco-mpc-BAD0}

\end{abstract}

\input{1_intro}

\input{2_model}
\input{3_mpc}

\input{4_rob_ada_mpc}
\input{5_bo_hyperparam}
\input{5_simu}

\input{6_real}
\input{7_conclusion}

\section*{Acknowledgments}
The authors would like to thank Johannes Köhler and Amon Lahr for their valuable discussions and feedback. This work was supported by NCCR Automation, grant agreement 51NF40\_225155 from the Swiss National Science Foundation, and by the European Union within the framework of the National Laboratory for Autonomous Systems (RRF-2.3.1-21-2022-00002).

\bibliography{reference}

\end{document}

%% file: 1_intro.tex
\section{Introduction}

\begin{figure}
    \centering
    \includegraphics[width=.95\linewidth]{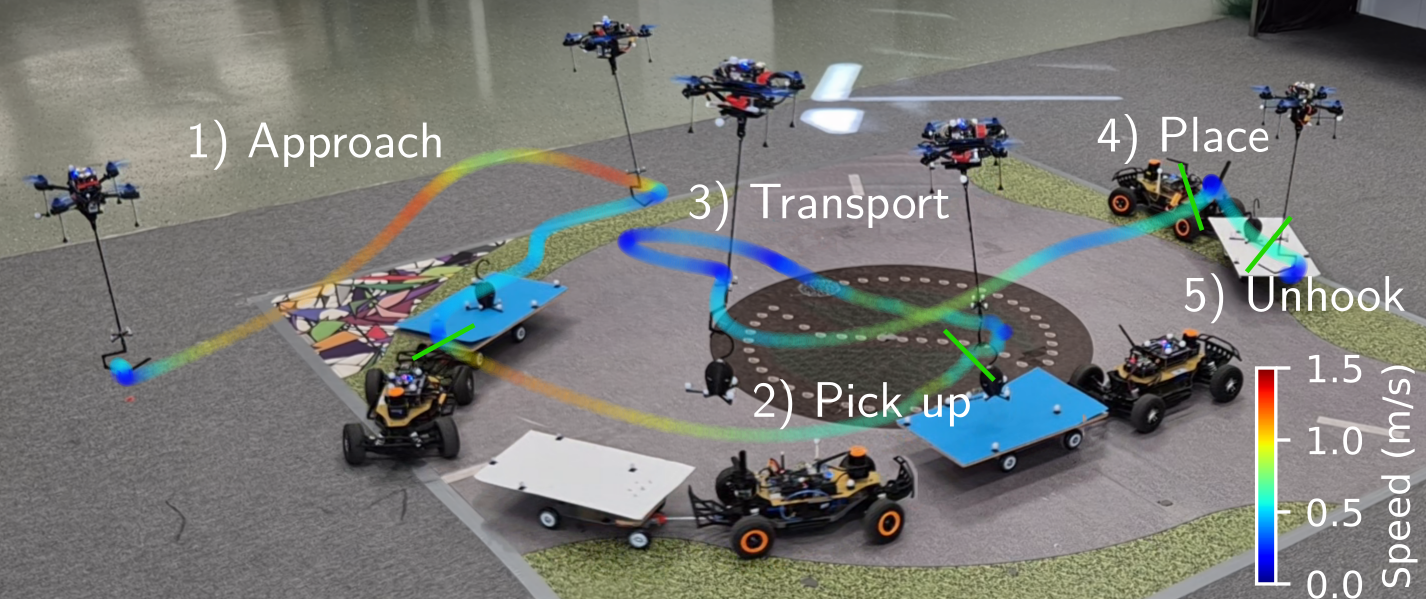}
    \vspace{-2mm}
    \caption{Dynamic pick-and-place with a hook-based aerial manipulator.}
    \label{fig:intro}
    \vspace{-5mm}
\end{figure}

Unmanned aerial vehicles (UAVs), and in particular 
quadrotors, are increasingly applied to solve complex tasks that require robotic manipulation and physical interaction with the environment, such as fixing high-voltage electric lines, cleaning windows, repairing blades of wind turbines, or package delivery \cite{Ruggiero2018}. In this work, we consider the challenging problem of dynamic pick-and-place between two moving ground platforms using a quadrotor equipped with a hook-based manipulator, as it is illustrated in Fig.~\ref{fig:intro}.

Aerial pick-and-place can be decomposed into three tasks: pick-up, transportation, and placement. To complete these tasks, two main configurations have become dominant: unmanned aerial manipulators (UAMs) by attaching robotic extensions to the quadcopter body \cite{Ollero2022, luo_time-optimal_2023, wang_impact-absorbing_2024, kumar_thrust-microstepping_2024}, and cable-suspended transportation systems \cite{Li2021, sun_agile_2025, li_autotrans_2023, wang_impact-aware_2024}. The full actuation and versatility of UAMs facilitate reliable and autonomous pick-and-place \cite{wang_impact-absorbing_2024, kumar_thrust-microstepping_2024}, but their relatively high weight and power consumption limit their flight time and maneuverability. 

Cable-suspended systems, by contrast, have recently demonstrated remarkable agility and precision. Early methods relied on geometric control, exploiting differential flatness \cite{Sreenath2013}, while current state-of-the-art approaches are based on nonlinear model predictive control (NMPC), with hybrid, adaptive, and learning-based extensions\cite{li_autotrans_2023, wang_impact-aware_2024, sarvaiya_hpa-mpc_2025, panetsos_gp-based_2024}.


Despite these advances, existing cable-based systems only partially address autonomous pick-and-place. Magnetic grippers have been used to pick up static payloads \cite{belkhale_model-based_2021}, but they require an external mechanism for drop-off. Autonomous placement onto moving platforms has been considered in \cite{yu_adaptive_2024}, but there the payload is pre-attached to the cable, so the controller does not need to solve the grasping problem. Hence, fully autonomous pick-and-place between moving platforms with a passive gripper has not yet been solved.

To combine the autonomous grasping ability of UAMs with the agility and energy efficiency of cable suspension, a hook-equipped aerial manipulator was introduced in \cite{antal2024autonomous} for static pick-and-place, and extended to dynamic grasping from a moving platform in \cite{antal_hook-based_2024}. In this design, a lightweight, passive hook is mounted on a rigid pole, which is connected to the quadrotor through an unactuated 2 DoF joint. While previous work on this platform demonstrated autonomous grasping and transportation, dynamic pick-and-place between two moving platforms has still remained a challenge.

In this work, we address these limitations by solving the following complex pick-and-place scenario. Initially, the payload with a hook on the top is carried by a moving ground platform. The quadrotor, equipped with a hook that connects to the drone through a rigid pole and 2 DoF joint, has to autonomously approach and grasp the payload with its hook, transport it to the target location, and place it onto another moving platform by detaching its hook, all within prescribed grasping and placement time windows, while satisfying state and input constraints under uncertain payload mass. 

From a control perspective, the introduced scenario 
poses several challenges. First, the coupled dynamics of the manipulator and the payload are highly nonlinear and include contact interactions that are difficult to capture accurately. Second, the system operates across multiple configurations (with and without attached payload), leading to abrupt changes in dynamic behavior. Third, uncertainty in the payload mass further affects the dynamics and must be handled in control design. Overall, achieving high performance requires sufficiently accurate models of these complex dynamics, and a control approach that effectively exploits these models. 

To provide an efficient solution for this problem, we propose a novel NMPC with the following contributions:
\begin{enumerate}[\setlength{\IEEEelabelindent}{0pt}
                   \setlength{\labelwidth}{0pt}
                   \setlength{\topsep}{0pt}
                   \setlength{\labelsep}{0.5em}]
    \item To capture the complex 
    dynamics of the transportation system, we use a MuJoCo-based \cite{todorov2012mujoco} digital twin directly as the predictive model of the NMPC, offering a simple, accurate, and easily reconfigurable approach.
    \item To handle the uncertainties of the predictive model due to uncertain payloads and external disturbances, we propose a robust adaptive NMPC that integrates zero-order uncertainty propagation with an extended Kalman filter for online parameter estimation, ensuring robust constraint satisfaction and high performance.
    \item We propose an optimization-based feasibility analysis to compute admissible pick-and-place time windows and certify constraint satisfaction over a set of scenarios, providing quantitative timing and safety guarantees.
    \item We demonstrate the effectiveness of the proposed approach through extensive simulation studies and real-world flight experiments for aerial pick-and-place between two moving ground platforms.
\end{enumerate}

%% file: 2_model.tex
\vspace{-1mm}
\section{MuJoCo model for hook-based pick-and-place}\label{sec:dyn}
\vspace{-1mm}


When describing the dynamics of an aerial transportation system, most works derive analytical models that rely on certain assumptions and simplifications, such as modeling the payload as a point mass, or assuming that the mass of the cable is negligible \cite{li_autotrans_2023, wang_impact-aware_2024, sarvaiya_hpa-mpc_2025}. In contrast, we use the physics simulator MuJoCo to generate a digital twin model of the hook-based system for three key reasons: i) it simplifies the modeling procedure, as instead of analytical derivations, the elements can be put together by joints of different characteristics;
ii) the resulting model is highly accurate, as the geometry and inertia of each component, including the hooks and the payload are directly considered; and iii) it is easy to reconfigure, as each element can be conveniently modified, if the design is changed (e.g. new hook geometry), and the corresponding model is auto-generated.

In modeling, we make one key simplification: to avoid discontinuities in the predictive model of the NMPC, we smoothly approximate the contact dynamics between the hook of the drone and the payload by augmenting the manipulator model with a 2 DoF revolute joint. Then, to address all phases of pick-and-place (with and without attached payload), we consider the mass of the payload as a parameter. During transportation, it equals to the actual mass value, otherwise it is set to zero to retrieve the dynamics of the manipulator. Later, we revisit the model mismatch corresponding to this approximation and integrate it into the robust MPC design, developed for flight control.

\begin{figure}
\vspace{2mm}
    \centering
    \includegraphics[width=0.4\linewidth]{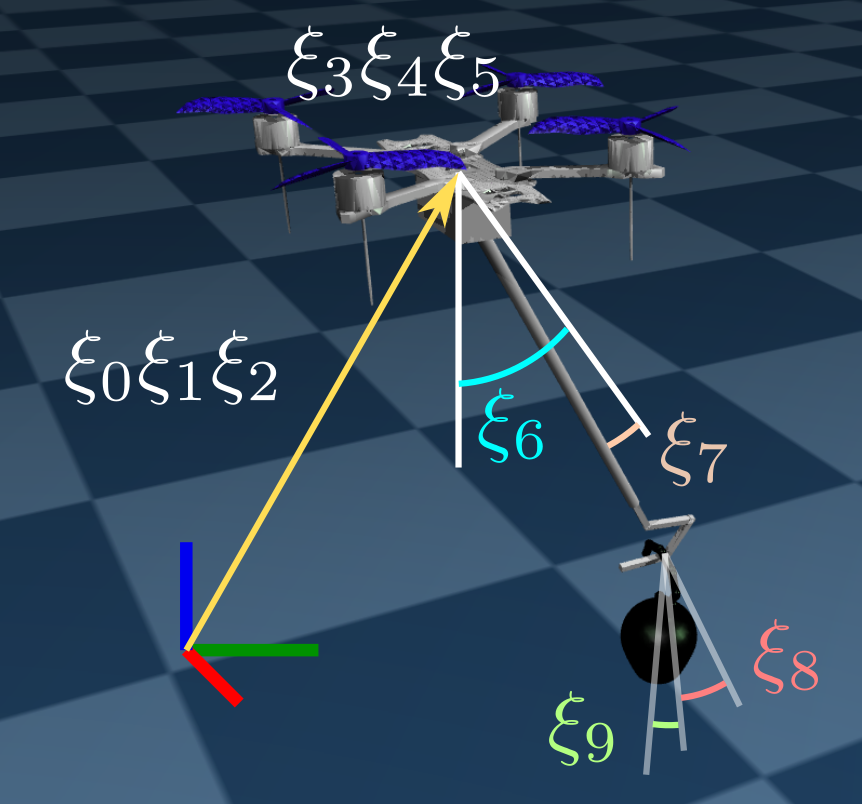}
    \vspace{-2mm}
    \caption{MuJoCo model describing the hook-based aerial manipulator and its coupled dynamics with a payload.}
    \label{fig:model}
    \vspace{-5mm}
\end{figure}

The resulting 10-DoF MuJoCo model with generalized coordinates $\xi_0$--$\xi_9$ is illustrated in Fig.~\ref{fig:model}. The quadcopter has 3 translational and 3 rotational DoFs. Then, the pole is connected through two passive revolute joints. Finally, two rotational DoFs represent the attachment of the hooks.


Formally, the dynamic model is described as follows:
{
\setlength{\abovedisplayskip}{4pt}
\setlength{\belowdisplayskip}{4pt}
\setlength{\abovedisplayshortskip}{2pt}
\setlength{\belowdisplayshortskip}{2pt}
\begin{align}\label{eq:dyn}
    \xi_{k+1} &= \phi(\xi_k, u_k, m_{\mathrm{L}, k}),
\end{align}
}%
where $\xi$ denotes the 20-dimensional discrete-time state vector with the first ten variables corresponding to the generalized coordinates, while the rest are their time derivatives, and $u = [\ F \ \tau^\top\ ]^\top$ is the control input vector containing the scalar collective thrust ($F$) and three-dimensional torque vector ($\tau$) around the quadcopter body-fixed axes. The discrete-time state transition function is denoted by $\phi$, the payload mass by $m_\mathrm{L}$, and $k$ is the discrete time index.

In this work, we assume that full state information is available for the controller. Measurements of the generalized coordinates can come from a motion capture system or encoders at the joints, while the velocity states can be either directly measured by inertial sensors or estimated by numerical differentiation of the position variables.




%% file: 3_mpc.tex
\vspace{-1mm}
\section{MPC design for dynamic pick-and-place}
\label{sec:mpc}
\vspace{-0.5mm}
\subsection{Overall concept}\label{sec:mpc:overall}
\vspace{-0.5mm}
Our control objective is to autonomously pick up a payload from a moving platform, transport it to another moving platform, and put it down. In addition, similar to 
\cite{antal_hook-based_2024}, we incorporate scheduling constraints similar to the 
time management in industrial processes. Specifically, the quadcopter has to grasp and put down the payload within dedicated time windows. In terms of grasping, we impose the constraint $\underline{T}_\mathrm{g} \leq T_\mathrm{g} \leq \overline{T}_\mathrm{g}$, and similarly for placement, $\underline{T}_\mathrm{p} \leq T_\mathrm{p} \leq \overline{T}_\mathrm{p}$. The beginning and end of the time windows ($\underline{T}_\mathrm{g}, \overline{T}_\mathrm{g}, \underline{T}_\mathrm{p}, \overline{T}_\mathrm{p}$) are fixed in the problem description, and the exact grasping and placement times ($T_\mathrm{g}, T_\mathrm{p}$) are obtained from the actual solution of the NMPC algorithm. In addition, we assume that the future motion of the moving platforms are known, e.g., obtained by forward simulating their digital twins in terms of MuJoCo models (as done in \cite{antal_hook-based_2024}).


In most of the relevant literature, NMPC is used for tracking, and a separate trajectory planner is developed (see e.g. \cite{li_autotrans_2023, wang_impact-aware_2024, sarvaiya_hpa-mpc_2025, panetsos_gp-based_2024}). In contrast, our proposed NMPC is used to design both the reference trajectory and the control inputs, providing a more holistic approach. Due to the fact that the goal is different in each phase of the task (grasping, transportation, and placement), we first define distinct motion phases with corresponding transition conditions. Then, we formulate a general optimal control problem (OCP), and describe the specific components of the OCP for each phase. 

We divide the pick-and-place task to five motion phases:
\begin{enumerate}[\setlength{\IEEEelabelindent}{0pt}
                   \setlength{\labelwidth}{0pt}
                   \setlength{\labelsep}{0.5em}]
    \item \textbf{Approach:} Get close to the payload until $t \geq \underline{T}_\mathrm{g}$
    \item \textbf{Pick up:} Attach the hook to the payload before $\overline{T}_\mathrm{g}$
    \item \textbf{Transport:} Move the load towards the target until  $t \!\geq\! \underline{T}_\mathrm{p}$
    \item \textbf{Place:} Put the payload on the target platform before $\overline{T}_\mathrm{p}$
    \item \textbf{Unhook:} Finish the task by detaching the hooks
\end{enumerate}

 



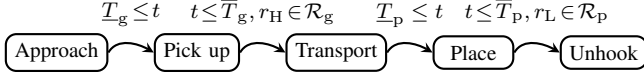
\begin{figure}
\vspace{2mm}
    \centering
    \input{phases_tikz}
    \vspace{-6mm}
    \caption{Task phases and transition conditions.}
    \label{fig:phases}
    \vspace{-5mm}
\end{figure}

The transitions between the phases are illustrated in Fig.~\ref{fig:phases}. Phase~2 is activated when the grasping time window specified by external scheduling opens, i.e., $k \geq \underline{N}_\mathrm{g}$, where $\underline{N}_\mathrm{g} = \lfloor \underline{T}_\mathrm{g}/\Delta t \rfloor$ with $\lfloor x \rfloor $ rounding to the nearest integer and $\Delta t$ being the discretization time step. 

To enter Phase~3, the hook needs to be attached. This condition is illustrated in Fig.~\ref{fig:mujoco_snapshot} and formulated as follows:
{
\setlength{\abovedisplayskip}{4pt}
\setlength{\belowdisplayskip}{4pt}
\begin{align}\label{eq:grasp_cond}
    r_{\mathrm{H}, k} \in \mathcal{R}_\mathrm{g}, \quad \mathcal{R}_\mathrm{g} = \{r\in \ril{3} \mid \| r - r_\mathrm{L,H}\|_2 \leq \rho_\mathrm{H} \},
\end{align}
}%
where $r_\mathrm{H}, r_\mathrm{L}$ are the position of the hook and the center of mass of the payload, $\rho_\mathrm{H}$ is the radius of the hook on the payload, and $r_\mathrm{L,H}$ is the position of the hook on the payload, computed as $r_\mathrm{L,H} = r_\mathrm{L} + R_\mathrm{L} e_3 d_\mathrm{L,H}$, where $R_\mathrm{L}$ is the rotation matrix of the payload and $e_3=[0\ 0\ 1]^\top$. 

Similar to Phase~2, Phase~4 starts when $k \!\geq\! \underline{N}_\mathrm{p}$ with $\underline{N}_\mathrm{p} =$ $ \lfloor \underline{T}_\mathrm{p}/\Delta t \rfloor$. Finally, Phase~5 is activated when the distance of the payload and the drop-off platform gets below a threshold:
{
\setlength{\abovedisplayskip}{4pt}
\setlength{\belowdisplayskip}{4pt}
\begin{align}\label{eq:dropoff_cond}
r_{\mathrm{L}, k} \in \mathcal{R}_\mathrm{p}, \quad \mathcal{R}_\mathrm{p} = \{r\in \ril{3} \mid \| r - r_\mathrm{p}\|_2 \leq \varepsilon_\mathrm{p} \},
\end{align}
}%
where $r_\mathrm{p}$ is the position of the drop-off platform and $\varepsilon_\mathrm{p}$ is a small constant, e.g., 1mm.

\vspace{-0.5mm}
\subsection{Optimization problem}
\vspace{-0.5mm}
We now introduce the generic form of the OCP, followed by the cost functions and constraints for each phase. 
Let $\boldsymbol{\xi} =  (\xi_{0 \mid k}, \dots, \xi_{N \mid k})$, $\boldsymbol{u}  =  (u_{0 \mid k}, \dots, u_{N-1 \mid k})$ be the predicted states and chosen inputs at time $k$. The optimal trajectories are computed as the solution of the discrete-time OCP:
{
\setlength{\abovedisplayskip}{4pt}
\setlength{\belowdisplayskip}{6pt}
\begin{subequations}\label{eq:mpc}
    \begin{align}
        \min_{\substack{\boldsymbol{\xi},\boldsymbol{u}}}  
        &\sum_{i=0}^{N-1} l\!\left(h_p(\xi_{i\mid k}), y_{p,k+i}^\mathrm{ref}\right) + \|v_{i\mid k} \|_{W_\mathrm{v}^p}^2 + \| u_{i\mid k} \|_{W_\mathrm{u}}^2 ,\label{eq:mpc:cost}
    \end{align}
    \vspace{-5mm}
    \begin{align}
        \text{s.t.} \quad 
        & \xi_{i+1\mid k} = \phi_p(\xi_{i\mid k}, u_{i\mid k}), 
        && \forall i\in\mathbb{I}_0^{N-1},\label{eq:mpc:dyn} \\
        & \xi_{0 \mid k} = \xi_k, \label{eq:mpc:init} \\
        & \underline{\xi} \leq \xi_{i+1\mid k} \leq \overline{\xi}, \ \ \underline{u} \leq u_{i\mid k} \leq \overline{u}, 
        && \forall i\in\mathbb{I}_0^{N-1},\label{eq:mpc:bounds:2} \\
        & g_p(\xi_{i\mid k}) \leq 0,
        && \forall i\in\mathbb{I}_0^{N}, \label{eq:mpc:nl_con}
    \end{align}
\end{subequations}
}%
where $h_p(\xi_{i\mid k}), y_{p,k+i}^\mathrm{ref}$ are the output and the output reference of the current phase, respectively, $v$ denotes the velocity states of $\xi$, $\| \cdot \|_{W}^2$ stands for the weighted Euclidean norm with positive definite weight matrix $W$, and $p \in \mathbb{I}_1^5$ represents the current motion phase. At each time $k$, $p$ is determined by the transition conditions of Fig.~\ref{fig:phases} and kept constant over the horizon. The dynamics, the initial conditions, and the state and input bounds are expressed by \eqref{eq:mpc:dyn}-\eqref{eq:mpc:bounds:2}, while \eqref{eq:mpc:nl_con} stands for nonlinear constraints on the state variables. The predictive model is set to $\phi_p(\cdot, \cdot)=\phi(\cdot, \cdot, 0)$ for $p\in\{1, 2, 5\}$ and $\phi_p(\cdot, \cdot)=\phi(\cdot, \cdot, m_\mathrm{L})$ otherwise. 

\begin{figure}
\vspace{2mm}
    \centering
    \includegraphics[width=.9\linewidth]{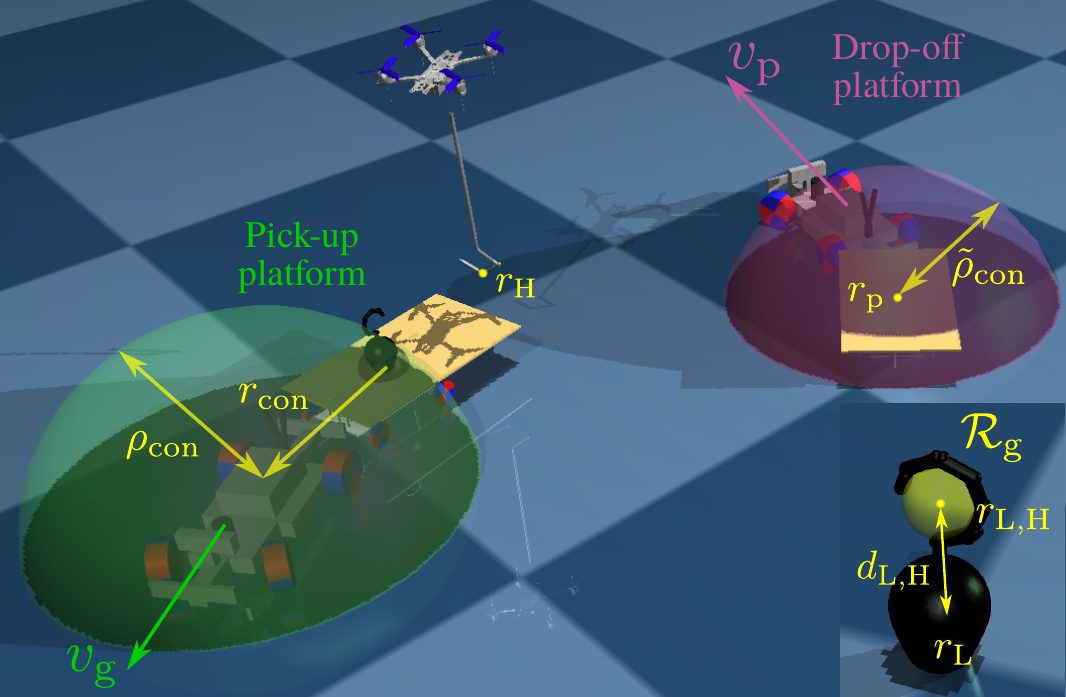}
    \vspace{-2mm}
    \caption{Snapshot from MuJoCo simulation of the dynamic pick-and-place task, illustrating the moving platforms, the quadrotor with hook, and the position, velocity, and geometric quantities used in the MPC formulation.}
    \vspace{-5mm}
    \label{fig:mujoco_snapshot}
\end{figure}

For stage cost, we use a smooth 1-norm approximation:
 $   l(y_p, y_p^\mathrm{ref}) = \sum_{j=0}^{n_\mathrm{y}-1}w_{p,j} \left( \sqrt{(y_{p,j} - y_{p,j}^\mathrm{ref})^2 + \gamma^2} - \gamma\right)$,
where $y_p, y_p^\mathrm{ref}$ are the output and output reference of phase $p$, and $w_{p,j}, \gamma >0$ are tuning parameters. 
Next, we introduce the specific cost terms and constraints for each phase.




\textbf{Phase 1:} To approach the payload, we penalize the difference of the position and orientation of the hook and the payload with the output and reference defined as follows:
\begin{align}
    h_1(\xi_{i\mid k})\! =\! \begin{bmatrix} r_{\mathrm{H},i\mid k}^\top \!\!& \psi_{\mathrm{H},i\mid k} \end{bmatrix}^\top\!\!\!\!,\ y_{1, k+i}^\mathrm{ref}\! =\! \begin{bmatrix} {r_{\mathrm{L, H}, k}^{\mathrm{ref}}}^\top \!\! & \psi_{\mathrm{L}, k}^{\mathrm{ref}} \end{bmatrix}^\top\!\!\!\!,\label{eq:cost_1}
\end{align}
where $\psi$ denotes the yaw angle of the hook and the payload. 
To make sure that the payload is not grasped for $k \leq \underline{N}_\mathrm{g}$, we introduce the following constraint:
\begin{align}
    g_1(\xi_{i\mid k}) \!=\! \rho_\mathrm{con} - \|r_{\mathrm{H},i\mid k} - (r_{\mathrm{L, H}, k+i}^{\mathrm{ref}} + R_{\mathrm{L}, k+i}^{\mathrm{ref}} r_\mathrm{con}) \|_2.\label{eq:pregrasp_cond}
\end{align}
Constraint \eqref{eq:pregrasp_cond} ensures that the position of the hook is outside of a sphere with radius $\rho_\mathrm{con}\in\mathbb{R}^+$ and center point $r_\mathrm{L}^{\mathrm{ref}} + R_\mathrm{L}^{\mathrm{ref}} r_\mathrm{con}$, where $r_\mathrm{con}$, $\rho_\mathrm{con}$ are tunable parameters. This is illustrated in Fig.~\ref{fig:mujoco_snapshot}, where the green shaded area represents the set that has to be avoided by the hook of the quadcopter.

\textbf{Phase 2:} Next, the objective is to grasp the payload, which is achieved by using the following output and reference:
\begin{align}
    h_2(\xi_{i\mid k}) \!=\! h_1(\xi_{i\mid k})
    ,\quad y_{2, k+i}^{\mathrm{ref}}\! = \!\begin{bmatrix} {r_{\mathrm{L, H}, k+i}^{\mathrm{ref}}}^\top\!\! & \psi_{\mathrm{L}, k+i}^{\mathrm{ref}} \end{bmatrix}^\top\!\!\!\!\!\!.
\end{align}
In this phase, nonlinear constraints are not imposed.

\textbf{Phase 3:} Once the hook is attached to the payload, the goal is to move the load to another moving vehicle (drop-off platform) with predicted position $r_\mathrm{p}^{\mathrm{ref}}$ and yaw angle $\psi_\mathrm{p}^{\mathrm{ref}}$. For this, the output and reference are defined as follows:
\begin{subequations}\label{eq:cost_3}
\begin{align}
        &h_3(\xi_{i\mid k}) = \begin{bmatrix} r_{\mathrm{L},i\mid k}^\top & \psi_{\mathrm{L},i\mid k} \end{bmatrix}^\top\!\!\!,\\ 
    & y_{3, k+i}^{\mathrm{ref}}  = \begin{bmatrix} (r_{\mathrm{p}, k+i}^{\mathrm{ref}}  + z_{\mathrm{safe}, k} e_3)^\top & \psi_{\mathrm{p}, k+i}^{\mathrm{ref}}  \end{bmatrix}^\top\!\!\!.\label{eq:cost_3:b}
\end{align}
\end{subequations}    
Notice that the payload position is now a decision variable, not an external signal, therefore the ${\mathrm{ref}} $ superscript is omitted. In \eqref{eq:cost_3:b}, $e_3 = [0 \ 0 \ 1]^\top$, and $z_{\mathrm{safe}, k} > 0$ is used to ensure that the payload is transported above the moving platform before putting it down. Its numerical value is calculated as follows: 
\begin{align}
    z_{\mathrm{safe}, k} = 0.5 \bar z \left(\mathrm{tanh}(a_1 d_{\mathrm{L,det}, k} - a_2)+1\right) \quad \forall k,
\end{align}
where $d_{\mathrm{L,det}, k} =\|(r_{\mathrm{L}, k} - r_{\mathrm{p}, k}^{\mathrm{ref}} )_{x, y} \|$ is the distance between the payload and the drop-off platform on the horizontal ($x,y$) plane, $\bar z \in \ril{+}$ is the reference height for transportation, and $a_1, a_2 \in \ril{+}$ are tunable parameters.

In this phase, 
similar to Phase 1, we introduce nonlinear constraints to prevent putting down the payload for $k \leq \underline{N}_\mathrm{p}$:
\begin{align}
    g_{3,m+1}(\xi_{i\mid k}) = \tilde\rho_\mathrm{con} - \|r_{\mathrm{L},i\mid k} - r_{\mathrm{p}, k+i}^{\mathrm{ref}}  \|_2.\label{eq:detach_cond}
\end{align}
Constraint \eqref{eq:detach_cond} ensures that the position of the transported payload is outside of a sphere with radius $\tilde\rho_\mathrm{con}\in\mathbb{R}^+$ centered around the target moving platform.

\textbf{Phase 4:} In this phase, the objective is similar to the previous one, however, instead of only moving towards the target, the aim is to put down the payload. Hence, the controlled output and the reference are the same as in Phase~3, $h_4(\xi_{i\mid k}) = h_3(\xi_{i\mid k})$, $y_{4, k+i}^{\mathrm{ref}}  = y_{3, k+i}^{\mathrm{ref}} $, but \eqref{eq:detach_cond} is no longer active, making the placement possible.

\textbf{Phase 5:} The objective of the last phase is to detach the hook from the payload:
\begin{subequations}\label{eq:cost_5}
\begin{align}    
    &h_5(\xi_{i\mid k}) = \begin{bmatrix} r_{\mathrm{H},i\mid k}^\top & \psi_{\mathrm{H},i\mid k} \end{bmatrix}^\top\!\!\!,\\ 
    & y_{5, k+i}^{\mathrm{ref}}  = \begin{bmatrix} (r_{\mathrm{p}, k+i}^{\mathrm{ref}}  + \bar x  R_{\mathrm{p}, k+i}^{\mathrm{ref}}  e_1)^\top & \psi_{\mathrm{p}, k+i}^{\mathrm{ref}}  \end{bmatrix}^\top\!\!\!,
\end{align}
\end{subequations} 
where $e_1 \! =\!  [1 \ 0 \ 0]^\top$ and $\bar x \! \in\! \ril{+}$ is a parameter that determines the setpoint of the hook. Its numerical value is chosen to be greater than the radius of the hook, e.g., $0.3$m.

%% file: phases_tikz.tex

\begin{tikzpicture}[
  box/.style={rectangle, draw, rounded corners, minimum height=1em, minimum width=3em, align=center, thick, font=\footnotesize},
  label/.style={font=\footnotesize}
]

\node[box] (approach) {Approach};
\node[box, right=0.6cm of approach] (pick) {Pick up};
\node[box, right=0.6cm of pick] (transport) {Transport};
\node[box, right=0.6cm of transport] (place) {Place};
\node[box, right=0.6cm of place] (unhook) {Unhook};

\draw[thick, ->, >=stealth] (approach.east) to[out=60,in=150] (pick.west);
\draw[thick, ->, >=stealth] (pick.east) to[out=60,in=150] (transport.west);
\draw[thick, ->, >=stealth] (transport.east) to[out=60,in=150] (place.west);
\draw[thick, ->, >=stealth] (place.east) to[out=60,in=150] (unhook.west);

\node[label, above=-0.3mm] at ($(approach.north east)!0.5!(pick.north west)$) {$\underline{T}_\mathrm{g}\! \le \! t$};

\node[label, above=0.0cm] at ($(pick.north east)!0.5!(transport.north west)$) {$t\! \leq\! \overline{T}_\mathrm{g}, r_\mathrm{H} \! \in \! \mathcal{R}_\mathrm{g}$};

\node[label, above=-0.3mm] at ($(transport.north east)!0.5!(place.north west)$) {$\underline{T}_\mathrm{p} \le t$};

\node[label, above=0.3mm]  at ($(place.north east)!0.5!(unhook.north west)$) {$t\! \leq\! \overline{T}_\mathrm{p}, r_\mathrm{L} \! \in \! \mathcal{R}_\mathrm{p}$};

\end{tikzpicture}

%% file: 4_rob_ada_mpc.tex
\section{Robust adaptive MPC (RAMPC)}\label{sec:rob_mpc}
\subsection{Problem formulation and uncertain dynamics}

In real-world aerial transportation, it is realistic to assume that the parameters of the manipulator are reasonably well known. However, there still remain significant model uncertainties, e.g. due to the contact dynamics between the hooks that we neglected in the predictive model (see Sec.~\ref{sec:dyn}), complex aerodynamics, and payloads with uncertain mass.

In this section, we introduce a robust adaptive MPC approach that is able to handle both parametric uncertainties and external disturbances. For this, first, we introduce the considered uncertain dynamic model. Then, we show how zero-order robust optimization (zoRO) \cite{zanelli_zero-order_2021} can be applied to solve the resulting OCP with minimal additional computational effort compared to the solution of \eqref{eq:mpc}. Finally, we describe how an extended Kalman filter (EKF) can be used to estimate uncertain parameters online and systematically improve the performance of the robust approach.

Let $\mathbb{S}_+^n$ denote the set of symmetric, positive definite matrices of dimension $n$. Then, for any $q\in\ril{n}$, $Q\in\mathbb{S}_+^n$, an ellipsoid is defined as 
 $ \ellips{q}{Q}\!  =\!  \{ x \! \in \! \ril{n} \mid  \|x-q\|_{Q^{-1}} \leq 1\}$.

We extend the predictive model \eqref{eq:mpc:dyn} with parametric uncertainty and external disturbance as follows:
\begin{align}\label{eq:unc_model}
    \xi_{k+1} \!=\! \phi (\xi_k, u_k, m_{\mathrm{L}, k}, \theta) + w_k \!=\! \phi_p (\xi_k, u_k, \theta) + w_k,
\end{align}
where $\phi_p(\cdot, \cdot, \cdot)\! =\! \phi(\cdot, \cdot, 0, \cdot)$ for $p\! \in\! \{1, 2, 5\}$ and $\phi_p(\cdot, \cdot, \cdot)\! =\! \phi(\cdot, \cdot, m_\mathrm{L}, \cdot)$ otherwise. Additionally, $\theta \! \in \! \ellips{\bar \theta}{W_\theta}$, $\bar\theta\! \in\! \ril{n_\theta}$, $W_\theta \in  \mathbb{S}_+^{n_\theta}$ is the vector of uncertain parameters, and $w_k  \in  \ellips{0}{W_\mathrm{w}}$,  $W_\mathrm{w} \in  \mathbb{S}_+^{n_\mathrm{w}}$ comprises unmodeled dynamics.

\subsection{ZoRO with ellipsoidal uncertainty sets}

Zero-order robust optimization (zoRO) \cite{zanelli_zero-order_2021, frey_efficient_2024} achieves computational efficiency by (i) approximating the true invariant set of the nonlinear system with ellipsoids and (ii) decoupling the computation of constraint tightening from the optimization, resulting in substantial dimensonality reduction. Next, based on \cite{zanelli_zero-order_2021, frey_efficient_2024}, we describe a robustified version of the OCP in \eqref{eq:mpc}, followed by the zoRO construction.

To extend the OCP formulation in \eqref{eq:mpc}, let us describe the predicted ellipsoidal uncertainty sets by $\boldsymbol{\Sigma} = (\Sigma_{0\mid k},\dots, \Sigma_{N \mid k})$, centered at $\bar{\boldsymbol{\xi}} =(\bar\xi_{0 \mid k}, \dots, \bar\xi_{N \mid k})$. Then, we formulate the following robustified OCP:
\begin{subequations}\label{eq:robust_mpc}
    \begin{align}
        &\!\!\!\!\!\!\!\!\!\!\! \min_{\substack{\bar{\boldsymbol{\xi}}, \boldsymbol{u}, \boldsymbol{\Sigma}}} 
        \sum_{i=0}^{N-1} l(\bar\xi_{i\mid k}, u_{i\mid k}),\\
        \text{s.t.} \ 
        & \bar\xi_{i+1\mid k} = \phi_p(\bar\xi_{i\mid k}, u_{i\mid k}, \bar \theta) ,
        && \forall i\in\mathbb{I}_0^{N-1}\\
        & \Sigma_{i+1\mid k} = \Phi(\Sigma_{i\mid k}, W, \bar\xi_{i\mid k}, u_{i\mid k}), 
        && \forall i\in\mathbb{I}_0^{N-1}\\
        & \bar\xi_{0 \mid k} = \xi_k, \quad \Sigma_{0 \mid k} = \bar\Sigma, \\
        & g(\bar\xi_{i\mid k}, u_{i\mid k}) + \beta(\bar\xi_{i\mid k}, u_{i\mid k}, \Sigma_{i\mid k})\leq 0,\!\!\!\!
        && \forall i\in\mathbb{I}_0^{N-1} 
    \end{align}
\end{subequations}
where $W=\mathrm{diag}(W_\theta,W_\mathrm{w})$, $g$ is the stacking of  \eqref{eq:mpc:bounds:2}-\eqref{eq:mpc:nl_con}, and $\bar\Sigma$ represents uncertainty in the measured state such that $\xi_k \in \ellips{\xi_{k}^\mathrm{true}}{\bar\Sigma}$. The backoff terms of the constraint tightening, denoted by $\beta$, are computed as follows:
\begin{align}
    \beta_{j}(\bar\xi_{i\mid k}&, u_{i\mid k}, \Sigma_{i\mid k}) =  \label{eq:backoff}\\ 
    &\sqrt{\nabla_\xi g_j(\bar\xi_{i\mid k}, u_{i\mid k})^\top \Sigma_{i\mid k} \nabla_\xi g_j(\bar\xi_{i\mid k}, u_{i\mid k})} \quad 
    \forall j \in \mathbb{I}_1^{n_\mathrm{g}}.\nonumber
\end{align}
The ellipsoidal uncertainty sets are propagated through the dynamics with the transition function $\Phi$ defined as follows:
\begin{align}
    \Phi(\Sigma, W, \xi, u)\!\! =\!\!   A(\xi, u) \Sigma A(\xi, u)\!^\top\!\! +\! G(\xi, u) W G(\xi, u)\!^\top\!\!,\label{eq:unc_prop}
\end{align}
where $A,G$ denote the Jacobians of the dynamics w.r.t the states and uncertainties:
\begin{align}
    A(\xi, u) = \frac{\partial \phi_p}{\partial \xi}\Big\vert_{\xi, u, \bar\theta},\  G(\xi, u) = \left[
        \frac{\partial \phi_p}{\partial \theta}\Big\vert_{\xi, u, \bar\theta} \ \ I_{n_\mathrm{w}}
    \right].
\end{align}
The uncertainty propagation \eqref{eq:unc_prop} is a standard approximation that has already proven to be a good tradeoff between accuracy and computational load \cite{zanelli_zero-order_2021}.

As it is explained in \cite{zanelli_zero-order_2021}, although \eqref{eq:robust_mpc} can be solved with standard nonlinear programming, the number of optimization variables is significantly increased compared to the nominal OCP given by \eqref{eq:mpc}. Hence, another approximation is introduced by computing the backoff terms using the previous solution and fixing them during the optimization. This way, the following zoRO problem is introduced:
\begin{subequations}\label{eq:zoro_mpc}
    \begin{align}
        &\!\!\!\!\!\!\!\!\!\!\!\!\!\!\min_{\substack{\bar{\boldsymbol{\xi}}, \boldsymbol{u}, \boldsymbol{\Sigma}}}  
        \sum_{i=0}^{N-1} l(\bar\xi_{i\mid k}, u_{i\mid k}),\\
        \text{s.t.}\ 
        & \bar\xi_{i+1\mid k} = \phi_p(\bar\xi_{i\mid k}, u_{i\mid k}, \bar \theta), 
        && \forall i\in\mathbb{I}_0^{N-1},\\
        & \bar\xi_{0 \mid k} = \xi_k,  \\
        &g(\bar\xi_{i\mid k}, u_{i\mid k}) + \hat\beta_i \leq 0 ,
        && \forall i\in\mathbb{I}_0^{N-1},\label{eq:zoro_mpc:constraints}
    \end{align}
\end{subequations}
where $\hat\beta_i$ are approximate backoff terms, recomputed at each iteration of the optimizer by (i) linearizing around the current nominal trajectory, (ii) propagating the uncertainties from initial condition $\bar\Sigma$ with \eqref{eq:unc_prop}, and (iii) applying \eqref{eq:backoff}.

\subsection{Extended Kalman filter for parameter estimation}\label{sec:ekf}



Robust MPC design ensures constraint satisfaction under uncertain parameters and bounded model mismatch. Next, we introduce an extended Kalman filter (EKF) for online parameter estimation to improve the tracking performance, estimate the uncertainty of the parameters, and reduce conservativeness. 
Later in the experiments, the EKF is used to estimate the mean value of the uncertain payload mass.


To adapt Kalman filtering for the task at hand, the uncertain dynamics \eqref{eq:unc_model} have to be revised. Specifically, we assume that the uncertainty $w_k \in \ellips{0}{W_\mathrm{w}}$ can be described by a normal distribution $\tilde w_k \sim \mathcal{N}(0, \Sigma_\mathrm{w})$. Then, we compute the covariance matrix based on the assumption that $\ellips{0}{W_\mathrm{w}}$ corresponds to the confidence region, as follows: 
\begin{align}\label{eq:bound_to_cov_ellips}
w_k^\top W_\mathrm{w}^{-1}  w_k \leq 1 \ \Leftrightarrow \ \tilde w_k^\top \Sigma_\mathrm{w}^{-1}  \tilde w_k \leq \chi_{n_\mathrm{w},\alpha}^2
\end{align}
where $\chi_{n_\mathrm{w},\alpha}^2$ is the value of the inverse CDF of the chi-squared distribution with $n_\mathrm{w}$ degrees of freedom for probability $\alpha$. From \eqref{eq:bound_to_cov_ellips}, $\Sigma_\mathrm{w}$ is expressed as
\begin{align}\label{eq:bound_to_cov}
    \Sigma_\mathrm{w} = W_\mathrm{w} / \chi_{n_\mathrm{w},\alpha}^2.
\end{align}
This way, the underlying model of the parameter estimator is formulated as follows:
    \begin{align}
    \bar\theta_{k+1} &= \bar\theta_{k} + \eta_{k}, \ \  y_k = d(\xi_{k}, u_k, \bar\theta, e_k)
    \label{eq:ekf_2}
\end{align}
where $\eta_{k} \sim \mathcal{N}(0, Q), Q \in \mathbb{S}_+^{n_\theta}$ is an artificial, small white noise input that controls the rate of the parameter change, while $e_k \sim \mathcal{N}(0, R), R \in \mathbb{S}_+^{n_\xi}$ models the measurement noise and the model uncertainty. The output is defined as
\begin{align}\label{eq:ekf_output}
\begin{split}
       y_k &= d(\xi_{k}, u_k, \bar\theta, e_k) = x_k + e_k \\
   &= \phi_p(\xi_{k-1}, u_{k-1}, \bar\theta) + \tilde w_{k-1} + e_k.
\end{split}
\end{align}
With the model and output equations in place, the parameter estimate $\hat\theta_k$ and the covariance $P_k$ can be computed using the standard EKF update equations, described e.g. in \cite{Plett2018}.
Note that EKF can only estimate constant or slowly varying parameters. If significant unmodeled dynamics are present, reliable adaptation requires recursive identification or learning methods that are computationally more expensive.

\subsection{Integration of EKF covariance and uncertainty bounds}

So far, we introduced two extensions to the nominal MPC: uncertainty propagation by zoRO and parameter estimation by EKF. 
Next, we derive a systematic integration of these two components by matching the covariance of the EKF estimate with the uncertainty bounds of the robust optimization.

In the robust OCP \eqref{eq:robust_mpc}, uncertain parameters are described by the ellipsoid $\theta \in \ellips{\bar \theta}{W_\theta}$. 
The estimated parameters by the EKF are tracked with a normal distribution $\theta_k \sim \mathcal{N}(\hat\theta_k, P_k)$. Similar to \eqref{eq:bound_to_cov_ellips}--\eqref{eq:bound_to_cov}, we match the confidence region of the normal distribution with the uncertainty bounds to arrive at the following update law:
\begin{align}\label{eq:ekf_zoro}
    W_{\theta, k} = \chi_{n_\theta, \alpha}^2 P_k.
\end{align}
By selecting the confidence level $\alpha$, the volume and reliability trade-off of the safety margin can be controlled by the user (e.g., 95\% is a standard choice).

Each time an EKF update is computed
, the current estimate $\hat\theta_k$ is used to evaluate the dynamics and the Jacobians in Optimization~\eqref{eq:zoro_mpc} and the covariance $P_k$ is used to update the uncertainty bound by \eqref{eq:ekf_zoro}, such that $W=\mathrm{blkdiag(W_{\theta, k}, W_\mathrm{w})}$. With this, we have a systematic integration of uncertainty propagation and parameter estimation.

%% file: 5_bo_hyperparam.tex
\section{Admissible time windows and feasibility analysis}\label{sec:bo_hyperparam}


In this section, we propose an optimization-based approach to analyze the feasibility of the proposed method over a bounded set of admissible scenarios. 
Specifically, we aim to determine \emph{admissible time windows}, parametrized by $\underline{T}_\mathrm{g}, \overline{T}_\mathrm{g}$, $\underline{T}_\mathrm{p}, \overline{T}_\mathrm{p}$, such that the RAMPC is feasible, all constraints are satisfied, and the pick-and-place is successful for every scenario in the considered sets.
To address this in a computationally tractable way, we make two assumptions: i) the task is executed in a fixed flight space with known bounds, and ii) a parametrization is available that describes the trajectories of the moving ground platforms. In automated logistics or manufacturing systems, these assumptions are realistic, since indoor spaces are bounded and conveyor belts or mobile robots move along well-defined paths.

\subsection{Admissible grasping time windows}\label{sec:bo_hyperparam:grasping}

We first provide an algorithm to compute admissible pick-up time windows $[\underline{T}_\mathrm{g}, \overline{T}_\mathrm{g}]$. The main challenges are that i) if $\underline{T}_\mathrm{g}$ is too small, the manipulator does not have enough time to start tracking the moving platform, and ii) if 
$\overline{T}_\mathrm{g} - \underline{T}_\mathrm{g}$ is too small, there is not enough time for grasping. 

Scenarios are characterized by a parameter vector $\eta$ that describes the initial position of the quadcopter and the trajectory of the moving platforms, and comes from a bounded set of admissible parameters $\mathcal{H}$. For a fixed time window $[\underline{T}_\mathrm{g}, \overline{T}_\mathrm{g}]$ and a given scenario $\eta$, 
we run closed-loop simulations with RAMPC and define the grasping time $T_\mathrm{g}(\eta, \underline{T}_\mathrm{g}, \overline{T}_\mathrm{g})$ as the first point in time for which the grasping condition \eqref{eq:grasp_cond} is satisfied. We say that RAMPC is \emph{feasible} for scenario $\eta$ and time window $[\underline{T}_\mathrm{g}, \overline{T}_\mathrm{g}]$ if it remains feasible at all time steps, all constraints are satisfied, and 
$\underline{T}_\mathrm{g} \leq  T_\mathrm{g}(\eta, \underline{T}_\mathrm{g}, \overline{T}_\mathrm{g})\leq \overline{T}_\mathrm{g}$. We call a window $[\underline{T}_\mathrm{g}, \overline{T}_\mathrm{g}]$ \emph{admissible} over $\mathcal{H}$ if RAMPC is feasible for all $\eta\in\mathcal{H}$.


To compute admissible grasping windows, we assume that the scenario set $\mathcal{H}$ and a user-defined upper bound $\overline{T}_\mathrm{g}^\mathrm{max}\! \! $ with $\overline{T}_\mathrm{g}\! \leq\! \overline{T}_\mathrm{g}^\mathrm{max}$ are available. As a first step, we set $\overline{T}_\mathrm{g}\! =\! \overline{T}_\mathrm{g}^\mathrm{max}\! \! $ and choose a $\underline{T}_\mathrm{g} \! =\!  \overline{T}_\mathrm{g}\!  -\!  \varepsilon_\mathrm{T}$, where $\varepsilon_\mathrm{T}$ is a small positive number (e.g., 0.1s). For this candidate window,  we run simulations for all $\eta\! \in\! \mathcal{H}$ and compute the \emph{worst-case grasping time}
\begin{align}\label{eq:tgmax}
    \overline{T}_\mathrm{g}^*=\max_{\eta\in\mathcal{H}} T_\mathrm{g}(\eta, \underline{T}_\mathrm{g}, \overline{T}_\mathrm{g}),
\end{align}
provided all scenarios remain feasible.

If there are infeasible scenarios (e.g., because $\overline{T}_\mathrm{g}\!  - \! \underline{T}_\mathrm{g}$ is too small), we decrease $\underline{T}_\mathrm{g}$ to enlarge the window until RAMPC is feasible for all $\eta\! \in\! \mathcal{H}$. If feasibility cannot be restored for any $\underline{T}_\mathrm{g} \! \leq \! \overline{T}_\mathrm{g}^\mathrm{max}\! \! \! $, then no admissible window exists for the given $(\overline{T}_\mathrm{g}^\mathrm{max}\! \! \! , \mathcal{H})$. In this case, the user must either increase $\overline{T}_\mathrm{g}^\mathrm{max}\! \! $ or restrict $\mathcal{H}$. Once feasibility is ensured, we use bisection to find the smallest feasible $\underline{T}_\mathrm{g}$. 
This smallest feasible lower bound is denoted by $\underline{T}_\mathrm{g}^*$, and the corresponding worst-case grasping time by $\overline{T}_\mathrm{g}^* = \max_{\eta\in\mathcal{H}} T_\mathrm{g}(\eta, \underline{T}_\mathrm{g}^*, \overline{T}_\mathrm{g}^\mathrm{max})$. 
As a result of the algorithm, we obtain that any admissible grasping window must satisfy $\underline{T}_\mathrm{g}^*\leq \underline{T}_\mathrm{g}$ and $\overline{T}_\mathrm{g}^* - \underline{T}_\mathrm{g}^* \leq \overline{T}_\mathrm{g} - \underline{T}_\mathrm{g}$, i.e., $\underline{T}_\mathrm{g}^*$ and $\overline{T}_\mathrm{g}^* - \underline{T}_\mathrm{g}^*$ are the lower bounds on the grasping start time and on the window width, respectively.


Solving \eqref{eq:tgmax} analytically would be extremely challenging, as each evaluation requires simulating a full scenario with RAMPC. Instead, we compute an approximation of the worst-case grasping time with Bayesian Optimization (BO) that systematically selects and evaluates scenarios in $\mathcal{H}$ to optimize \eqref{eq:tgmax} for a given candidate window.

To additionally verify constraint satisfaction for a given window, we compute the \emph{worst-case constraint violation}
{
\begin{align}\label{eq:nugmax}
    \nu_\mathrm{g}^*=\max_{\eta\in {\mathcal{H}}} \max_{k\in \mathbb{I}_0^{\overline{N}_{\mathrm g}} } g(\xi_k^\eta, u_k^\eta),
\end{align}
}%
where $g$ is the constraint vector from \eqref{eq:zoro_mpc:constraints} and $(\xi_k^\eta, u_k^\eta)$ denotes the closed-loop trajectory for scenario $\eta$. If $\nu_\mathrm{g}^* \leq 0$, the window $[\underline{T}_\mathrm{g}, \overline{T}_\mathrm{g}]$ is admissible over $\mathcal{H}$. Otherwise, the scenario set $\mathcal{H}$ or the parameters of the RAMPC must be adjusted. The worst-case value of $\nu_\mathrm{g}^*$ is approximately computed by BO, analogous to the solution of \eqref{eq:tgmax}.


\vspace{-1mm}
\subsection{Admissible placement time windows}\label{sec:bo_hyperparam:placement}
\vspace{-1mm}

An analogous procedure is used to determine admissible placement windows $[\underline{T}_\mathrm{p}, \overline{T}_\mathrm{p}]$. Here, the goal is to ensure that the payload can be transported from the pick-up configuration to the drop-off platform and placed within the time interval, for a set of scenarios $\mathcal{M}$. Scenarios are characterized by a parameter vector $\mu$ that includes the initial configuration, the trajectories of the moving platforms, and the payload mass. 
For fixed 
$[\underline{T}_\mathrm{p}, \overline{T}_\mathrm{p}]$ and $\mu$, we run closed-loop simulations with RAMPC (assuming the payload is already attached) and define the drop-off time $T_\mathrm{p}(\mu, \underline{T}_\mathrm{p}, \overline{T}_\mathrm{p})$ as the first point in time for which the drop-off condition \eqref{eq:dropoff_cond} is satisfied. We use the same terminology as before: RAMPC is feasible for $\mu$ and  $[\underline{T}_\mathrm{p}, \overline{T}_\mathrm{p}]$ if it remains feasible, satisfies all constraints, and achieves $\underline{T}_\mathrm{p}\! \leq\!   T_\mathrm{p}(\mu,\!  \underline{T}_\mathrm{p},\!  \overline{T}_\mathrm{p})\! \leq\!  \overline{T}_\mathrm{p}$; the window is admissible over $\mathcal{M}$ if this holds for all $\mu\! \in\! \mathcal{M}$.

From preliminary simulations, 
we observe that successful placement does not require tracking of the target platform, hence, we fix $\underline{T}_\mathrm{p}\! =\! 0$ and search for the smallest admissible $\overline{T}_\mathrm{p}$. 
For this, we specify a user-defined upper bound $\overline{T}_\mathrm{p}^\mathrm{max}$ 
with $\overline{T}_\mathrm{p} \! \leq \! \overline{T}_\mathrm{p}^\mathrm{max}\! $ and compute the \emph{worst-case drop-off time}
{
\begin{align}\label{eq:tpmax}
    \overline{T}_\mathrm{p}^*=\max_{\mu\in\mathcal{M}} T_\mathrm{p}(\mu, 0, \overline{T}_\mathrm{p}^\mathrm{max}),
\end{align}
}%
using BO. Finally, to certify constraint satisfaction and feasibility for all $\mu\in\mathcal{M}$ with a candidate $\overline{T}_\mathrm{p}^*$, we compute the worst-case constraint violation in the transportation phase
{
\begin{align}\label{eq:nupmax}
    \nu_\mathrm{p}^*=\max_{\mu\in {\mathcal{M}}} \max_{k\in \mathbb{I}_0^{\overline{N}_{\mathrm p}} } g(\xi_k^\mu, u_k^\mu),
\end{align}
}%
where the notations are analogous to \eqref{eq:nugmax}. If $\nu_\mathrm{p}^* \leq 0$, the placement window $[0, \overline{T}_\mathrm{p}^*]$ is admissible over $\mathcal{M}$, otherwise either $\mathcal{M}$ or the RAMPC parameters must be adjusted. Finally, as a result of the algorithm, we obtain a lower bound $\overline{T}_\mathrm{p}^*$ on the width of the placement time window.


%% file: 5_simu.tex
\vspace{-0.5mm}
\section{Simulation study}\label{sec:simu}
\vspace{-0.5mm}

\subsection{Test environment}
\vspace{-0.5mm}

In our test arena, the available flight space is a $7\times 5 \times 3\ \mathrm{m}$ rectangle. To carry the payload, we use F1Tenth ground vehicles \cite{OKelly20_f1tenth} with custom designed trailers. For the simulation study, the digital twin model of each component has been implemented in MuJoCo as illustrated in Fig.~\ref{fig:mujoco_snapshot}. Throughout the analyzed scenarios in this paper, the ground platforms are moving along paperclip-shaped trajectories with constant speed. These trajectories are characterized by two parameters: the starting point along this periodic path ($s_\mathrm{g}\in[0, 1]$) and the speed of the towing vehicle ($v_\mathrm{g}$ for the pick-up platform, $v_\mathrm{p}$ for the drop-off platform).

We run the MPC on a desktop PC with Intel i9 CPU. For the numerical solution, the SQP-RTI solver of \texttt{acados} is used \cite{verschueren_acadosmodular_2022}. The Jacobians of the MuJoCo dynamics are computed with finite differences, 
and the predictive model is discretized with the \emph{implicit-in-velocity Euler} integrator of MuJoCo and with sampling time $\Delta t = 0.05$s. The state measurement uncertainty bound is set to $\bar\Sigma = 10^{-4}I_{n_{\xi}}$. The horizon length is $N = 25$, and the solution is recomputed with a control frequency of 50Hz. In simulation, this proven to be sufficient, and, in real experiments, the computed body rates are sent to the quadcopter and tracked by a low-level PD controller that runs onboard at 500Hz. To solve \eqref{eq:tgmax}--\eqref{eq:nupmax} numerically, we use BoTorch \cite{balandat2020botorch}. 

As explained in Sec.~\ref{sec:dyn}, the predictive model used in the NMPC relies on a smoothened hook–payload interaction. In contrast, the main simulation loop uses the internal contact model of MuJoCo. 
This induces a deliberate mismatch between the predictive and simulation models, so that the robustness of the proposed controller can be evaluated against more realistic contact dynamics.

\subsection{Admissible time windows}

\begin{figure}
\vspace{2mm}
    \centering
    \includegraphics[width=\linewidth]{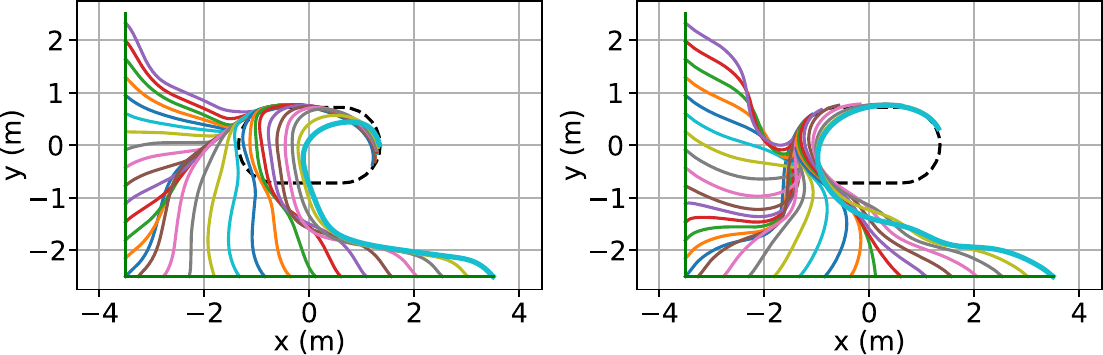}
    \vspace{-7mm}
    \caption{$(x, y)$ position trajectories of the quadcopter before grasping the payload (left) and during transportation (right). The black dashed line shows the path of the moving platform, while the green line shows the set of initial positions when solving \eqref{eq:tgmax} and \eqref{eq:tpmax}. The bold cyan curve highlights the worst-case trajectory in terms of pick-up and drop-off time, respectively.}
    \label{fig:bo_grasp}
    \vspace{-5mm}
\end{figure}

Next, the time window parameters are optimized to make sure the pick-and-place will be successful under the available flight space and moving platform trajectory. The scenario parameter vector is defined as $\eta = [p_\mathrm{xy}\ s_\mathrm{g}\ v_\mathrm{g}] \in \mathbb{R}^3$. Here, $p_\mathrm{xy} \in [0, 1]$ is introduced to describe the set of initial positions for the quadcopter with a single scalar variable. The $(x,y)$ initial positions are fixed to the boundary of the flight space, shown by the green lines in Fig.~\ref{fig:bo_grasp}, to reduce the dimensionality of $\eta$, and the initial $z$ coordinate is fixed to 1m. The mapping $p_\mathrm{xy}\rightarrow(x,y)$ is defined as follows:
{
\setlength{\abovedisplayskip}{4pt}
\setlength{\belowdisplayskip}{4pt}
\setlength{\abovedisplayshortskip}{2pt}
\setlength{\belowdisplayshortskip}{2pt}
\begin{align*}
    (x, y) \!=\! \begin{cases}
        (-3.5, 2.5 - 10 p_\mathrm{xy}),\!\! & \text{if }p_\mathrm{xy} \in[0, 0.5]\\  
        (-3.5 + 14 (p_\mathrm{xy} - 0.5), -2.5), \!\!\!\!& \text{if } p_\mathrm{xy} \in (0.5, 1].
    \end{cases}
\end{align*}
}%
%
Further, $s_\mathrm{g} \in [0, 1]$ admits the moving platform to start from any point of the periodic path, and $v_\mathrm{g}\in [0.4, 0.6]\mathrm{m/s}$ is the permitted speed of the moving platform.


The computed worst-case admissible grasping time windows, see Sec.~\ref{sec:bo_hyperparam:grasping}, are $\underline{T}_\mathrm{g}^* = 4.3$s and ${T}_\mathrm{g}^*= 7.7$s, with $\eta^* = [1.0\ 0.0\ 0.6]$. 
This means that if the start of the grasping time window is $\underline{T}_\mathrm{g} \geq 4.3$s and its duration is $\overline{T}_\mathrm{g} - \underline{T}_\mathrm{g} \geq 3.4$s, then the manipulator has enough time to start tracking the moving platform in Phase~1 and successfully pick up the payload in Phase~2. Simulated trajectories of the quadcopter are displayed in Fig.~\ref{fig:bo_grasp}, showing that $\underline{T}_\mathrm{g} = 4.3$s is indeed enough to start tracking the moving payload from any initial position along the boundaries of the flight space. The worst-case trajectory is indicated by the bold cyan curve.

Next, we look for the feasible set of time windows for the transportation and dynamic placement of a payload with uncertain mass. In this case, the scenario parameter vector is  $\mu = [p_\mathrm{xy}\ s_\mathrm{p}\ v_\mathrm{p}\ m_\mathrm{L}] \in \mathbb{R}^4$. The admissible sets of the first three parameters are the same as previously, and the payload mass is restricted to $m_\mathrm{L}\in[0.05, 0.2]\mathrm{kg}$. Its nominal value (used to initialize the EKF) is fixed to $m_\mathrm{L}^0= 0.07\mathrm{kg}$.

The numerical result of solving \eqref{eq:tpmax} with BO is ${T}_\mathrm{p}^* = 10.28\mathrm{s}$ with $\mu^* = [1.0\ 0.29\ 0.6\ 0.05]$. 
This means that the worst-case time needed to transport the payload to the target platform and put it down under the considered assumptions is $10.28\mathrm{s}$. The right panel of Fig.~\ref{fig:bo_grasp} shows simulated trajectories of the quadcopter during transportation, with the worst-case scenario $\mu^*, {T}_\mathrm{p}^*$ highlighted by the bold cyan curve.

With this, we have successfully computed the worst-case time window parameters under the considered assumptions, and have shown that even for these scenarios, the dynamic pick-and-place is always accomplished. 

\addtolength{\tabcolsep}{-2pt}
\begin{table}[b]
\vspace{-4mm}
    \centering
    \caption{Performance of nominal and RAMPC in simulation.
    }\vspace{-2mm}
\begin{tabular}{lcccccccc}
\hline
Mass deviation & \multicolumn{2}{c}{0\%} & \multicolumn{2}{c}{10\%} & \multicolumn{2}{c}{20\%} & \multicolumn{2}{c}{50\%} \\
Controller & Nom & RA & Nom & RA & Nom & RA & Nom & RA \\
Success rate (\%) & 92 & 100 & 91 & 100 & 86 & 100 & 39 & 100 \\
Average cost & 4.98 & 6.47 & 5.17 & 6.53 & 5.19 & 6.54 & 5.16 & 6.63 \\
\hline
\end{tabular}
    \label{tab:simu_performance}
\end{table}
\addtolength{\tabcolsep}{2pt}

\vspace{-1mm}
\subsection{Comparison of nominal and RAMPC in simulations}
\vspace{-1mm}

To show the benefits of the robust adaptive extension, we now provide a comparison with the nominal MPC that is described in Sec.~\ref{sec:mpc}. For this, we run various simulations by randomizing the scenario parameters $\mu = [p_\mathrm{xy}\ s_\mathrm{g}\ v_\mathrm{g}\ m_\mathrm{L}]$. Moreover, to evaluate the performance of the controllers under a mismatch between the nominal and actual payload mass values, we simulate all scenarios with $0\%$,$\pm 10\%$, $\pm 20\%$, and $\pm 50\%$ difference between these values. The results of simulating 800 scenarios overall are shown in Table~\ref{tab:simu_performance}. The proposed RAMPC achieves a 100\% success rate in all cases. In contrast, the nominal controller loses feasibility in some scenarios even when the nominal and actual payload mass coincide, resulting in 92\% success rate. With increasing payload mass deviation, the success rate slightly decreases, until it significantly drops at 50\% deviation. 

Looking at the average costs, we see that the cost achieved by RAMPC is around 25\% higher than the baseline. This is due to the conservativeness of the 
uncertainty propagation of zoRO. However, this performance tradeoff is justified by the 100\% success rate over all examined scenarios.

The computation times of the nominal and RAMPC are displayed in the right panel of Fig.~\ref{fig:simu_robustness}, showing that the proposed approach requires only an additional 1.5ms (both in average and worst-case) compared to the baseline, making it real-time feasible at 50Hz by a significant margin.

\addtolength{\tabcolsep}{-2pt}
\begin{table}[b]
    \vspace{-4mm}
    \centering
    \caption{Scenario parameters for the experiments.}\vspace{-2mm}
    \begin{tabular}{l|cccccccc}
         & $x_\mathrm{0}$ & $y_\mathrm{0}$ & $v_\mathrm{g,p}$&$ \underline{T}_\mathrm{g}$&$ \overline{T}_\mathrm{g}$ & $\underline{T}_\mathrm{p}$&$ \overline{T}_\mathrm{p}$ & $m_\mathrm{L}$ \\
         \hline
        \#1 & 2.0 & -1.0 & 0.4 & 12.0 & 15.5 & 16.0 & 23.0 & 0.2\\
        \#2 & -2.0 & -1.0 & 0.5 & 6.0 & 10.0 & 16.0 & 22.0 & 0.1\\
        \#3 & -2.0 & 1.0 & 0.6 & 9.5 & 13.5 & 13.5 & 19.0 & 0.15\\
    \end{tabular}
    \label{tab:exp_params}
\end{table}
\addtolength{\tabcolsep}{2pt}

\vspace{-1mm}
\subsection{Representative scenarios}
\vspace{-1mm}

\begin{figure}
\vspace{2mm}
    \centering
    \includegraphics[width=.9\linewidth]{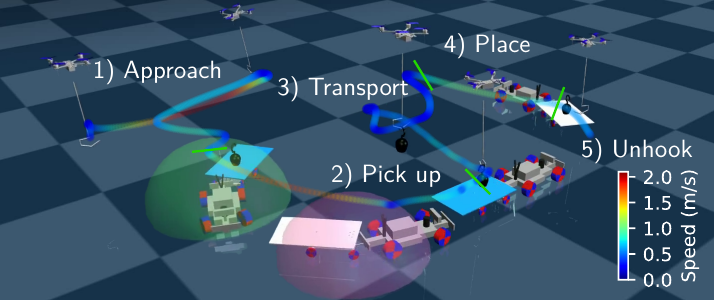}
    \vspace{-2mm}
    \caption{Dynamic pick-and-place in simulation by RAMPC. The trajectory of the hook of the aerial manipulator is displayed in colors, while the boundaries of the motion phases are indicated with green dashes.}
    \label{fig:simu_trajs}
\end{figure}

\begin{figure}
    \centering
    \includegraphics[scale=0.62]{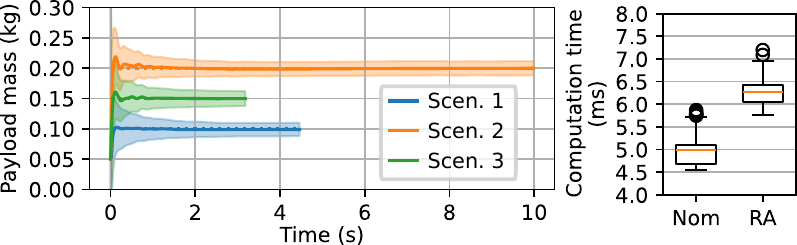}
        \vspace{-4mm}
    \caption{Left: payload mass estimation by RAMPC in simulations, with $\pm3\sigma$ interval corresponding to the EKF covariance. The horizontal axis shows $t\in[0, T_\mathrm{p}-T_\mathrm{g}]$. Right: computation time of nominal and RAMPC.}
    \label{fig:simu_robustness}
    \vspace{-5mm}
\end{figure}

We further evaluate the performance of RAMPC with three dedicated scenarios, the parameters of which are provided in Table~\ref{tab:exp_params}. The initial position of the quadrotor, speed of the moving platforms, the time windows and the mass of the payload are varied across the scenarios. The EKF is initialized by $m_\mathrm{L}^0 = 0.05\mathrm{kg}$ for all scenarios. As a result of the simulations, the pick-up and drop-off times are $(T_\mathrm{g}, T_\mathrm{p}) = [(13.52, 18.98), (8.08, 19.06), (11.52, 15.72)]\mathrm{s}$, all within their respective bounds given in Table~\ref{tab:exp_params}. The left panel of Fig.~\ref{fig:simu_robustness} shows the convergence of the EKF in these simulated scenarios (in the respective $[T_\mathrm{g}, T_\mathrm{p}]$ intervals). The estimator converges rapidly to the actual mass value (indicated with dashed lines) and remains precise for the entire transportation phase. The motion trajectory of Scenario~2 is shown in Fig.~\ref{fig:simu_trajs} with snapshots from the MuJoCo simulation. 



%% file: 6_real.tex
\vspace{-0.5mm}
\section{Flight experiments}
\vspace{-0.5mm}

We evaluate the real-world performance of the RAMPC in flight experiments. Our quadrotor is equipped with a Crazyflie Bolt flight controller\footnote{\url{https://www.bitcraze.io/products/crazyflie-bolt-1-1/}} that receives body rate commands from the PC and runs a low-level PD controller at 500Hz, using the on-board IMU measurements. Optitrack 
is used to provide high-precision pose information.

\begin{figure}
\vspace{2mm}
    \centering
    \includegraphics[scale=.62]{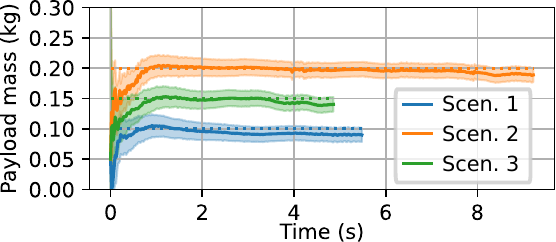}
    \vspace{-2mm}
    \caption{Payload mass estimation by RAMPC in flight experiments, with $\pm3\sigma$ interval corresponding to the EKF covariance. The horizontal axis shows $t\in[0, T_\mathrm{p}-T_\mathrm{g}]$, where the EKF is active.}
    \label{fig:param_estimation_real}
    \vspace{-5mm}
\end{figure}

We investigate the three scenarios that are specified by the parameter values in Table~\ref{tab:exp_params}. Similar to the simulation study, the EKF is initialized by $m_\mathrm{L}^0 = 0.05\mathrm{kg}$ for all scenarios. The actual pick-up and drop-off times in these experiments are $(T_\mathrm{g}, T_\mathrm{p}) = [(14.12, 20.62), (7.92, 18.16), (11.06, 16.94)]\mathrm{s}$, all within their respective bounds, similar to the simulations. The online parameter estimation by the EKF is displayed in Fig.~\ref{fig:param_estimation_real} (in the respective $[0, T_\mathrm{p}- T_\mathrm{g}]$ intervals). Although the convergence of the estimator is a bit slower than in simulations ($\sim$1s), the estimation always stays within the confidence interval $\pm 3\sigma$, and after convergence, it remains very close to the actual value. 

Fig.~\ref{fig:intro} shows the measured trajectory of the hook of the quadrotor 
in Scenario~2. The motion phases can be clearly identified: the quadcopter starts tracking the pick-up platform and then maintains the speed $v_\mathrm{g}\!=\!0.5\mathrm{m/s}$. Once Phase~2 starts, it accelerates to attach the hooks. Next, it starts tracking the drop-off platform, and finally puts down the payload by detaching the hooks. 
Recordings of all three representative scenarios are included in the supplementary video.



%% file: 7_conclusion.tex
\vspace{-2mm}
\section{Conclusion}\label{sec:conclusion}
\vspace{-0mm}

This paper presented a robust adaptive nonlinear MPC approach for autonomous pick-and-place between moving platforms using a hook-equipped aerial manipulator. 
By integrating zero-order robust optimization and EKF-based online parameter estimation into a digital twin-based MPC, robust constraint satisfaction and high performance under uncertain payload mass is achieved. An optimization-based feasibility analysis further computes admissible grasping and placement time windows and provides safety guarantees. Extensive simulations and real-world flight experiments demonstrate safe and precise payload transportation with consistent performance across significant payload mass variations.

Future work aims to solve fully onboard deployment, e.g. using vision-based sensing to eliminate reliance on external sensors, and extending the approach to outdoor settings. 